\documentclass[letterpaper, 10 pt, conference]{ieeeconf}  

\IEEEoverridecommandlockouts                              

\usepackage{graphicx} 
\graphicspath{ {images/} }
\usepackage{algorithm}
\usepackage[noend]{algpseudocode}
\usepackage{float}
\usepackage{comment}
\usepackage{amsmath} 
\usepackage{amssymb}  
\DeclareMathOperator*{\argmax}{argmax}
\usepackage{hyperref}

\title{\LARGE \bf
Preemptive Motion Planning for \\ Human-to-Robot Indirect Placement Handovers
}

\author{Andrew Choi$^{1}$, Mohammad Khalid Jawed$^{2}$, and Jungseock Joo$^{3}$
\thanks{*This work was supported by the National Science Foundation (Grant No. IIS-1925360).}
\thanks{$^{1}$Department of Computer Science,
        University of California, Los Angeles
        {\tt\small asjchoi@cs.ucla.edu}}%
\thanks{$^{2}$Department of Mechanical and Aerospace Engineering,
        University of California, Los Angeles
        {\tt\small khalidjm@seas.ucla.edu}}%
\thanks{$^{3}$Department of Communication,
        University of California, Los Angeles
        {\tt\small jjoo@comm.ucla.edu}}%
\thanks{Further info can be found at \url{pmp-human-to-robot.github.io}}%
\thanks{To appear in IEEE ICRA 2022}%
}

\begin{document}

\maketitle
\thispagestyle{empty}
\pagestyle{empty}

\begin{abstract}
As technology advances, the need for safe, efficient, and collaborative human-robot-teams has become increasingly important. One of the most fundamental collaborative tasks in any setting is the object handover. Human-to-robot handovers can take either of two approaches: (1) direct hand-to-hand or (2) indirect hand-to-placement-to-pick-up. The latter approach ensures minimal contact between the human and robot but can also result in increased idle time due to having to wait for the object to first be placed down on a surface. To minimize such idle time, the robot must preemptively predict the human intent of where the object will be placed. Furthermore, for the robot to preemptively act in any sort of productive manner, predictions and motion planning must occur in real-time. We introduce a novel prediction-planning pipeline that allows the robot to preemptively move towards the human agent's intended placement location using gaze and gestures as model inputs. In this paper, we investigate the performance and drawbacks of our early intent predictor-planner as well as the practical benefits of using such a pipeline through a human-robot case study.

\end{abstract}

\section{INTRODUCTION}
Industrial robots have traditionally been utilized in caged or taped-off settings for the safety of nearby human operators. These robots, often ``hard-coded", are assigned highly specified and repetitive tasks requiring a large amount of precision. As technology advances, an increase in the need for more intelligent robots has been seen as robots are applied to more dynamic, human-populated settings such as warehouses, restaurants, and public streets. One of the most challenging aspects of integrating robots into everyday life is in developing systems that can safely work with humans with minimal human supervision. In order to ensure safe and fluid operation, robots must be able to infer human intent and factor in such information into their actions accordingly, much like real human workers.

One of the most fundamental tasks between workers in a team environment is the handover of tools or materials. Direct mid-air hand-to-hand transfers require the robot to move in a way that avoids collision with the human operator while making contact with the held out object. This requires accurate estimation of the transfer point \cite{nemlekar2019objecttransferpoint} as well as grasp classification \cite{yang2020grasp}.
Indirect placement transfers (i.e. placing an object down so that the opposite agent can pick it up easily) circumvent this problem entirely albeit with reduced efficiency due to the extra step of having to place the object down. Still, indirect placement transfers are heavily prevalent all throughout daily life such as when dealing with dangerous tools (passing knives) or transferring objects to occupied individuals (placing empty glasses at a bar for a bartender). Therefore, with accurate prediction of the object placement intent along with preemptive planning, robots can take advantage of the benefits of indirect handovers while minimizing inherent inefficiencies.

\begin{figure}
  \centering
  \includegraphics[width=3.3in]{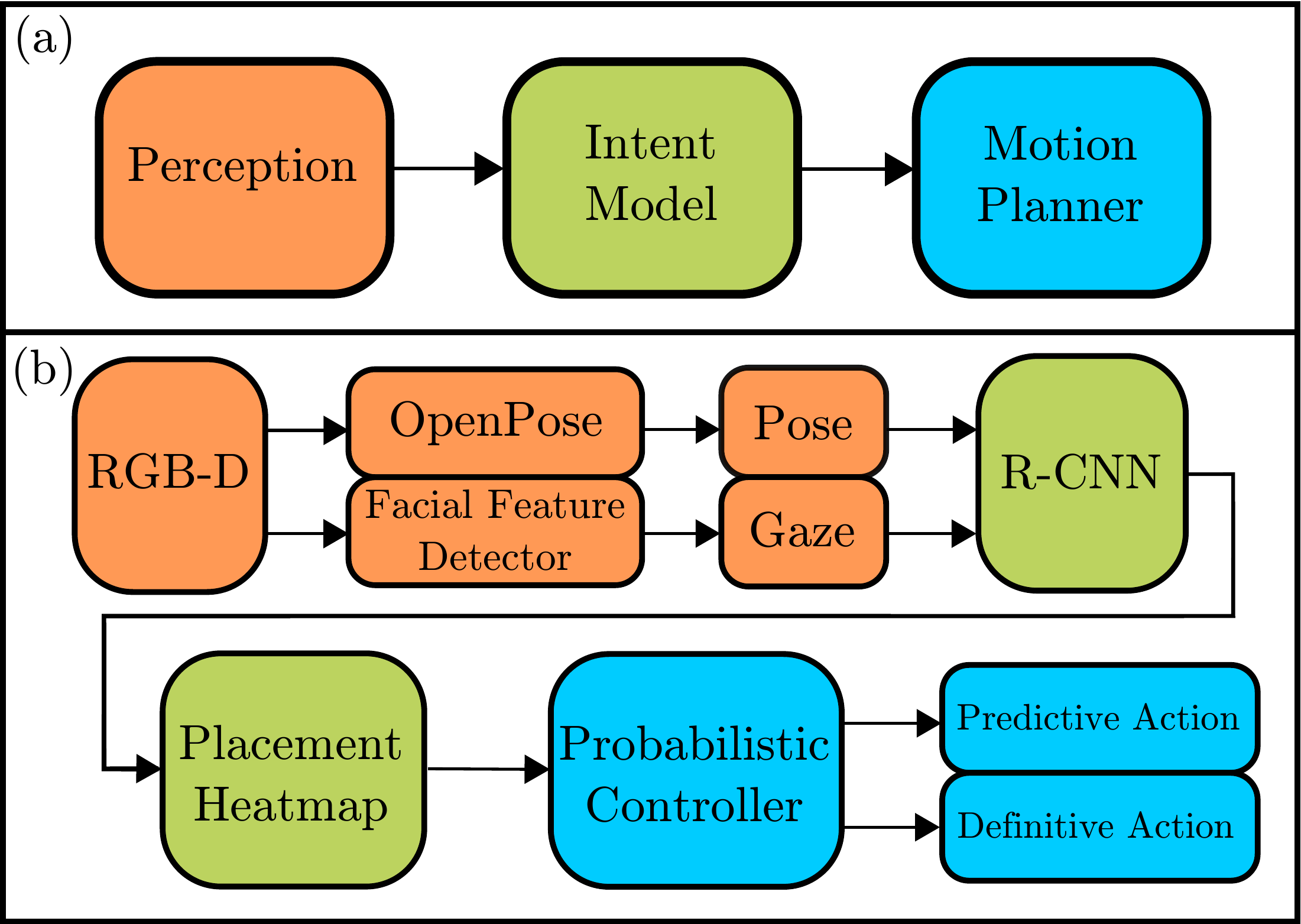}
  \caption{Robot system pipeline. (a) shows the high level modules of the system which are color coded. (b) shows the individual components making up each of the high level modules. Predictive and definitive actions are defined in Sec. \ref{subsec:motion_planning}.}
  \label{fig:pipeline}
\end{figure}

\begin{figure*}
  \centering
  \includegraphics[width=\textwidth]{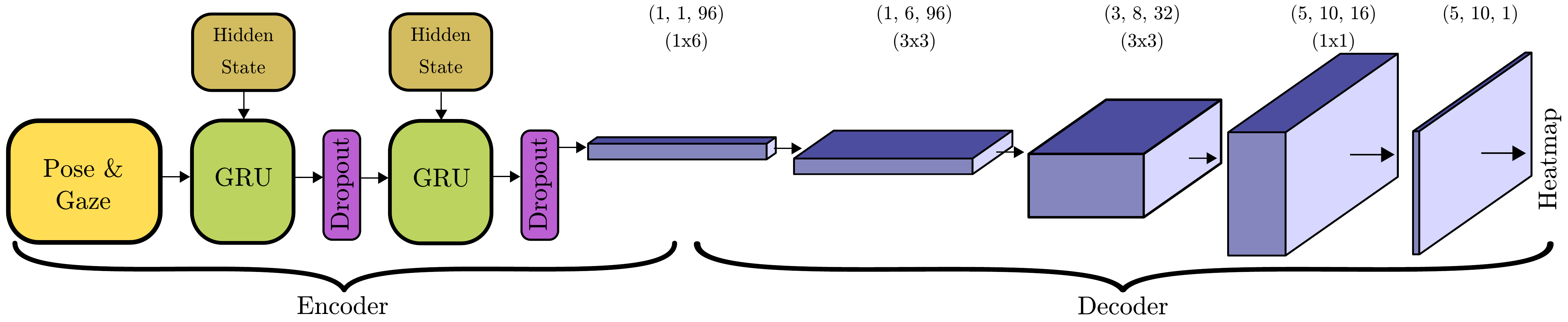}
  \caption{An illustration of the intent model architecture. As shown, the model is a recurrent convolutional network. Recurrent units act as the encoder, where the pose \& gaze feature vector is fed into the gated recurrent units (GRU) at every time step. Here, the hidden states fed back into the GRUs represent the encoded state of past time steps. Convolutional layers act as the decoder which perform a series of transpose convolutions from a 1-D encoded state vector. The dimensions of the convolutional layers are shown for a heatmap output of size $5 \times 10$, which is the size used in the case study.}
  \label{fig:network_architecture}
\end{figure*}

In this paper, we investigate the problem of preemptive motion planning for efficient human-to-robot indirect object placement transfers using real-time predictions of human object placement intent as feedback. We introduce a prediction-planning pipeline that allows the robot to preemptively move towards objects placed by a human. The primary contributions for our work are as follows: (1) we introduce a prediction-planning pipeline for preemptive human robot placement handovers that can serve as a basis for future system implementations; (2) we show practical quantitative results in a human-robot-collaboration case study utilizing preemptive motion planning when compared to more traditional wait-and-act approaches; (3) we analyze human social perception of preemptive robots; and (4) we show that accurate real-time performance can be achieved without expensive prohibitive equipment such as motion capture or gaze tracking eyewear.

The remainder of the paper is as follows. In Sec. \ref{sec:related_work}, we review prior work on human intent prediction and preemptive control. In Sec. \ref{sec:methodology}, we discuss the overall methodology involving the intent model and how predictions are factored into the robot's motions. Next, in Sec. \ref{sec:experimental_design}, we go over the experiment design while in Sec. \ref{sec:evaluation}, we analyze the results from the case study. Finally, conclusive remarks as well as potential future research directions are discussed in Sec. \ref{sec:conclusion}.

\section{RELATED WORK}
\label{sec:related_work}

Human intent prediction has been a popular research topic tackled by a variety of communities in machine learning, computer vision, and robotics~\cite{doshi2009roles,razin2017learning}. Intent prediction has especially become a point of interest for the robotics community as humans become increasingly integrated into robot workspaces. Such inference is necessary for a variety of human-robot-collaborative tasks such as load sharing co-manipulation \cite{Townsend2017EstimatingHI, Tariq2018loadsharing}, socially-aware navigation \cite{chen2017social, gupta2018social}, and object placement handovers \cite{Chen2018trust, kratzer2020anticipating, bansal2020supportive}, which commonly adopt non-verbal cues such as gesture, gaze, and facial expression to predict human intention or attention~\cite{park2013predicting,nehaniv2005methodological,joo2014visual,fan2019understanding,admoni2017social,saran2018human,kratzer2020mogaze}.

In terms of the social and cognitive aspects of the human-robot handover task, previous research has focused on understanding human preference of handovers \cite{cakmak2011preferences, quispe2017preference}, the effect of gaze \cite{moon2014meetgaze, admoni2014gazedelays, schydlo2018rnn_anticipation, kshirsagar2020gazebehaviors}, human comfort \cite{pan2018socialperception}, as well as the roles of trust \cite{walker2015trust, Chen2018trust} and support \cite{bansal2020supportive}. Attempts to develop frameworks that codify the coordination structure of handovers have also been made \cite{strabala2013seamless}. 

At a lower level, previous methods for human intent prediction have utilized Gaussian mixture models (GMM) to learn a library of human motions for early prediction of human reaching motions \cite{mainprice2013earlyprediction}. Later work showed quantitative results showcasing the benefits of utilizing GMM produced early predictions in tasks with humans and robots in the same workspace \cite{luo2018unsupervised}.


More recently, deep learning methods have also been used for intent prediction. Recurrent neural networks (RNN) have been shown to be able to classify high level human actions using gaze and body cues \cite{schydlo2018rnn_anticipation}. More similar to our specific problem, RNNs have also been shown capable of predicting object placement intents when used with Mixture Density Networks \cite{kratzer2020anticipating}.
Other methods have attempted to use inverse reinforcement learning approaches in order to obtain a cost function explaining human behavior from human demonstrations \cite{mainprice2015irl, chen2017social, sun2018irl}, or to learn how to interpret human intent directly from interaction and communication without demonstrations~\cite{wu2021communicative}.

Following this, the robotics community has utilized such human intent prediction methods in order to develop preemptive control methods. Such prior work has worked on developing robot systems that are capable of preemptively planning with human intent factored in \cite{luo2018unsupervised, huang2016anticipatory, jain2020anticipatory}. Similar to these works, we introduce a novel preemptive control method using data-driven intent prediction that is specifically catered towards human-to-robot object placement handovers. Furthermore, we also investigate the impact of preemptive planning on human comfort and preference. 

\begin{figure*}
  \centering
  \includegraphics[width=0.91\textwidth]{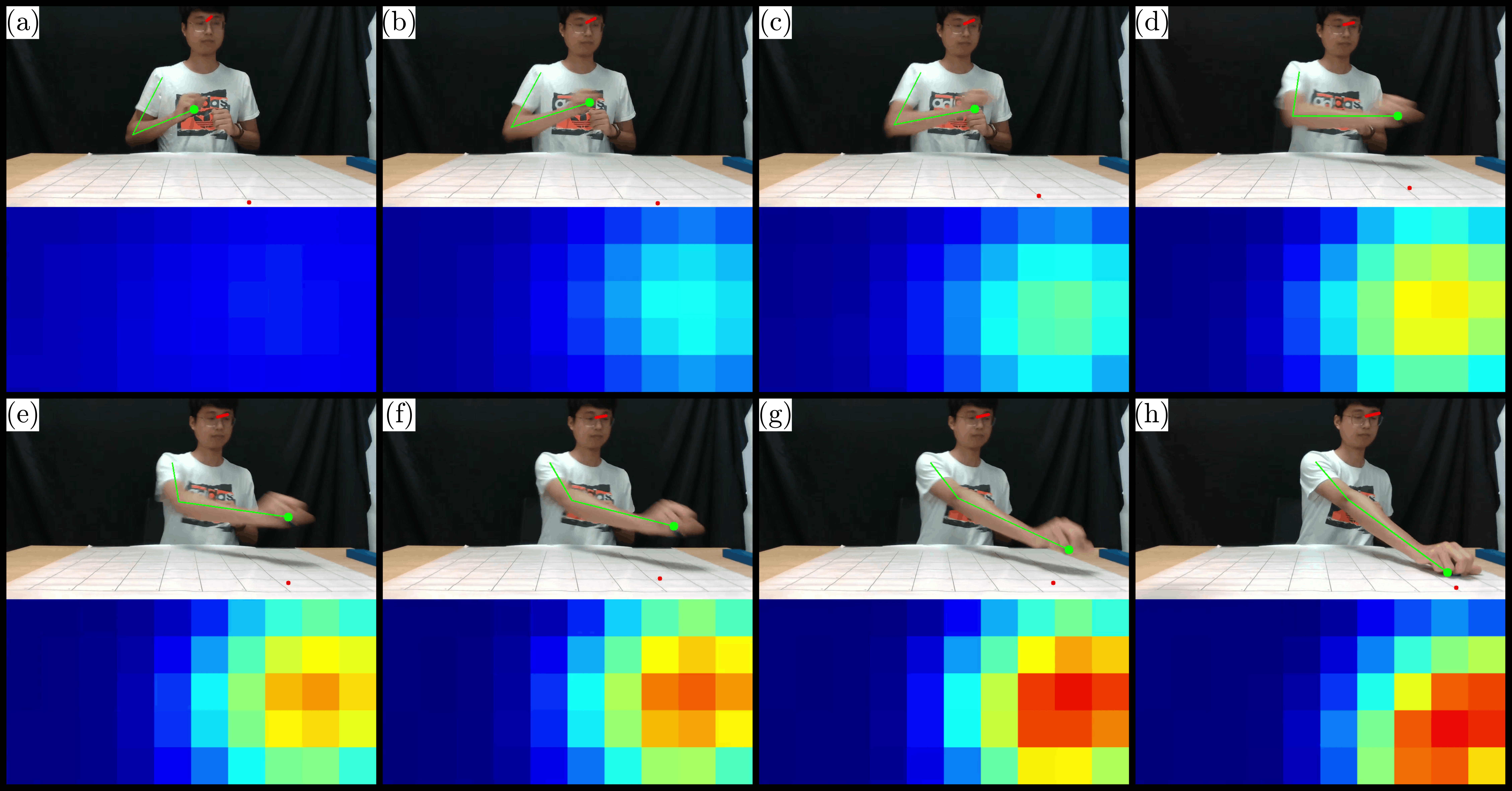}
\vspace{-5pt}
  \caption{An example prediction sequence. The red vector located on the participant's head is the detected face norm. The red dot indicates the gaze and table plane intersection point $\boldsymbol \psi$ discussed in Sec. \ref{subsec:features}. The green skeleton shows the detected shoulder, elbow, and palm positions. For each frame (a-h) along the trajectory, the corresponding heatmap can be seen underneath. Note that at the start of the trajectory, the model is unsure of the human's placement intent. As more and more of trajectory is realized, a prediction with increasing confidence can be seen being made.}
  \label{fig:prediction_example}
\end{figure*}

\section{METHODOLOGY}
\label{sec:methodology}

\subsection{Intent Prediction}
For the predictive model, we train a recurrent convolutional network as shown in Fig. \ref{fig:network_architecture}.
This model takes in features constructed from human gesture and gaze in order to produce a $n \times m$ heatmap representing the human agent's object placement likelihoods on an opposing flat surface. We incorporate both pose and gaze as inputs as a combination of both has been shown to produce better predictive results than one alone \cite{schydlo2018rnn_anticipation}.

\subsection{Features}
\label{subsec:features}

For the first set of features, we simply obtain the 3D positions of the palm, elbow, and shoulder which are obtained through Openpose \cite{cao2019openpose, joshi2019rosopenpose}. The second set of features involve the head pose and norm. To avoid the complexities of pupil tracking, we use facial feature detection \cite{lemaignan2016realtime} to obtain the head pose $\mathbf h$ as well as the concurrent reference frame where the head norm is the x-axis.

For the next feature, we compute the intersection point $\boldsymbol \psi$ of the human's gaze vector with the table plane.
To obtain this, we first tilt the head norm vector 30 degrees along the y-axis to point towards the table and treat this subsequent vector $\mathbf g$ as the gaze vector. Then, to obtain the gaze intersection point along the table plane, we compute
\begin{equation}
    \boldsymbol \psi = \mathbf h - \left( \frac{\mathbf n_\textrm{t} \cdot (\mathbf h - \mathbf p)} {\mathbf n_\textrm{t} \cdot \mathbf g} \right) \mathbf g,
\end{equation}
where $\mathbf n_\textrm{t}$ is the norm of the table plane and $\mathbf p$ is an arbitrary point on the table plane. As we normalize the table plane to be through the origin, we can remove the $z$-component of $\boldsymbol \psi$ so that $\boldsymbol \psi$ consists of only the $x$ and $y$ coordinates. These features as well as their velocities computed from the previous set of features round out the inputs to our model.

\subsection{Labels}
To train the model, we use a $n \times m$ grid of probabilities as our labels. A sigmoid layer is used as our output layer to avoid near zero probability values as the grid size increases. Furthermore, to avoid overconfident predictions resultant of one-hot encoding, we use a 2D Gaussian to employ label smoothing for a label 
\begin{equation}
    l(x,y) = \frac{1}{2 \pi s_x s_y} \exp \left(-\frac{(x-m_x)^2}{2 s_x^2} - \frac{(y-m_y)^2}{2 s_y^2} \right),
\end{equation}
where $x \in [0, n-1]$ and $y \in [0, m-1]$ are the label indices with respect to the grid, $m_x$ and $m_y$ are the true positions, and $s_x$ and $s_y$ are the standard deviations for the Gaussian. 

Additionally, we also scale the labels by weights 
\begin{equation}
    c_t = 1 - \exp \left( -\frac{5t}{T} \right),
\label{eq:weights}
\end{equation}
where $t$ is the current time step, and $T$ is the total length of the trajectory. This is done so that the model is initially unsure and increases its confidence as more of the trajectory is observed. To properly apply the confidence weight, we normalize the labels so that the maximum value is 1 to obtain our final label
\begin{equation}
    l_t(x,y) = \frac{l(x,y)}{\max(\mathbf L)} c_t,
\end{equation}
where $\mathbf L$ is the set of all labels. 

\subsection{Loss}
Similar to the labels, we assign the weights from Eq. \ref{eq:weights} to our training loss as well so that earlier incorrect predictions are penalized less than later ones. As we are working with a probability regression problem, we employ mean squared error (MSE) as our loss function. The loss for one training trajectory can then be computed as
\begin{equation}
    loss = \sum_{t=0}^{T-1} \left( \sum_{y=0}^{m-1} \sum_{x=0}^{n-1} \text{MSE}(l_t(x, y), p_t(x, y) ) \right) c_t,
\end{equation}
where $p_t(x, y)$ is the prediction of the $x$-th and $j$-th grid location.

\subsection{Weighted Predictions}
\label{subsec:weighted_pred}
Finally, when obtaining predictions during real-time operation, we keep the last $h$ predictions in order to obtained weighted grid probabilities. Similar to \cite{huang2016anticipatory}, we use a set of weights
\begin{equation}
    w_i = (1 - \epsilon)^i \ \forall \ i \in [h],
\end{equation}
where $\epsilon$ is a decay factor. Applying these weights allow for more recent predictions to have exponentially more influence. The weighted prediction for a certain grid location and current time step can then be computed as
\begin{equation}
    \bar{p}(x,y) = \frac{1}{h} \sum_{i=0}^{h-1} p_i(x,y) w_i.
\end{equation}

\subsection{Preemptive Motion Planning}
\label{subsec:motion_planning}

\begin{figure*}
  \centering
  \includegraphics[width=0.95\textwidth]{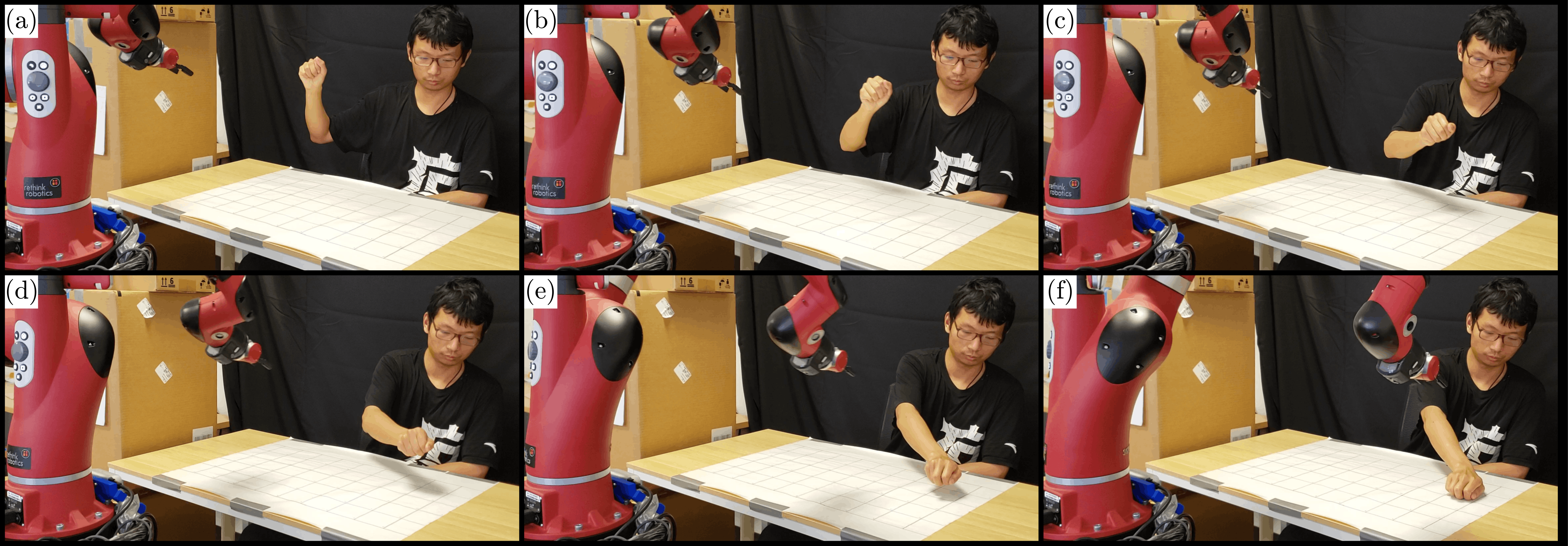}
  \caption{Example sequence with preemptive motion planning. The robot is in the ready position in frame (a). As more and more of the trajectory is observed, the robot correctly predicts the human's placement intent and is able to move to a position to pick up the object as shown in frame (f).}
  \label{fig:motion_planning_example}
\end{figure*}

To carry out motion planning, we employ stochastic trajectory optimization for motion planning (STOMP) \cite{kalakrishnan2011stomp} as this algorithm has been shown to be capable of producing smooth trajectories in real-time \cite{mainprice2013earlyprediction}.
At every time step, we calculate the full weighted heatmap $\bar{\mathbf p}$ which can then be used to obtain the grid location containing the highest probability $\hat{p} = \max (\bar{\mathbf p})$.
If $\hat{p}$ surpasses an execution limit $\gamma$ and the final object location is still unknown, we have the robot carry out a \textit{predictive action} towards the grid. Otherwise, if the final object location becomes known, we carry out a \textit{definitive action} to pick up the object.

Two separate threads are used to asynchronously carry out predictive and definitive actions based on the current known state. 
If a predictive action is already in progress when the final object position becomes known, we preempt and replan a definitive action. If an ongoing predictive action is close enough (\textit{we used a tolerance of 1 x-grid and 2 y-grids}) to the known goal position, we carry out the full motion in order to minimize any stop-and-go stuttering. 
The predictive and definitive algorithms are shown in Algs. \ref{alg:predictive_alg} and \ref{alg:definitive_alg}, respectively.

\section{Experimental Design}
\label{sec:experimental_design}
To analyze the benefits of using our prediction-planning pipeline, we perform a comparative case study for the human-to-robot object placement transfer task. 
For this task, the human places down wooden cubes measuring 1 inch on all sides on a table for the robot to pickup and sort. In an effort to minimize idle time and maximize efficiency, the robot will attempt to preemptively move to where the human places the object before the final object position becomes known. 

\begin{algorithm}
\caption{Predictive Node}
\begin{algorithmic}
\State memory $\gets$ queue of size $h \times n \times m$ initialized to all zeros
\While{running and no object is detected}
    \State $\mathbf s \gets$ obtain skeleton from Openpose
    \State $\mathbf g \gets$ obtain gaze from facial feature detection
    \State $\mathbf f \gets$ concatenate $\mathbf s$ and $\mathbf g$
    \State $\mathbf p \gets$ forward pass with $\mathbf f$
    \State memory.push($\mathbf p$)
    \State $\bar{\mathbf p} \gets$ obtain weighted predictions from Sec. \ref{subsec:weighted_pred}
    \State $\hat{p} \gets \max (\bar{\mathbf p})$ 
    \If{$\hat{p} > \gamma$}
        \State $(x, y) \gets \argmax (\bar{\mathbf p})$ \Comment{our current guess}
        \If{no predictive action in progress}
            \State carry out \textit{predictive action} towards $(x, y)$
        \ElsIf{previous goal is not within tolerance}
            \State preempt and carry out new \textit{predictive action}
        \EndIf
    \EndIf
\EndWhile
\end{algorithmic}
\label{alg:predictive_alg}
\end{algorithm}

\begin{algorithm}
\caption{Definitive Node}
\begin{algorithmic}
\While{running}
    \State object position $\gets$ detection algorithm
    \If{object is detected}
        \If{no predictive action in progress}
            \State carry out \textit{definitive action}
        \ElsIf{previous goal not within tolerance}
            \State preempt and carry out \textit{definitive action}
        \EndIf
    \EndIf
\EndWhile
\end{algorithmic}
\label{alg:definitive_alg}
\end{algorithm}

\subsection{Baseline}
For our baseline, we create a naive planning pipeline that works purely reactively. In other words, the robot does not take into consideration any human-related factors and simply plans to grab the cube once the cube is detected. We will call this approach \textit{reactive} and our proposed method as \textit{preemptive}. Both methods were set to use identical joint speeds for fair comparison and consisted of three sequential motions: x-y traversal, move to pre-grasp position, and move to grasp position. More sophisticated baselines such as following the arm trajectory were forgone due to an excessive amount of stop-and-go stuttering caused by constant motion replanning (something our preemptive approach avoids).

\begin{figure*}
  \centering
  \includegraphics[width=\textwidth]{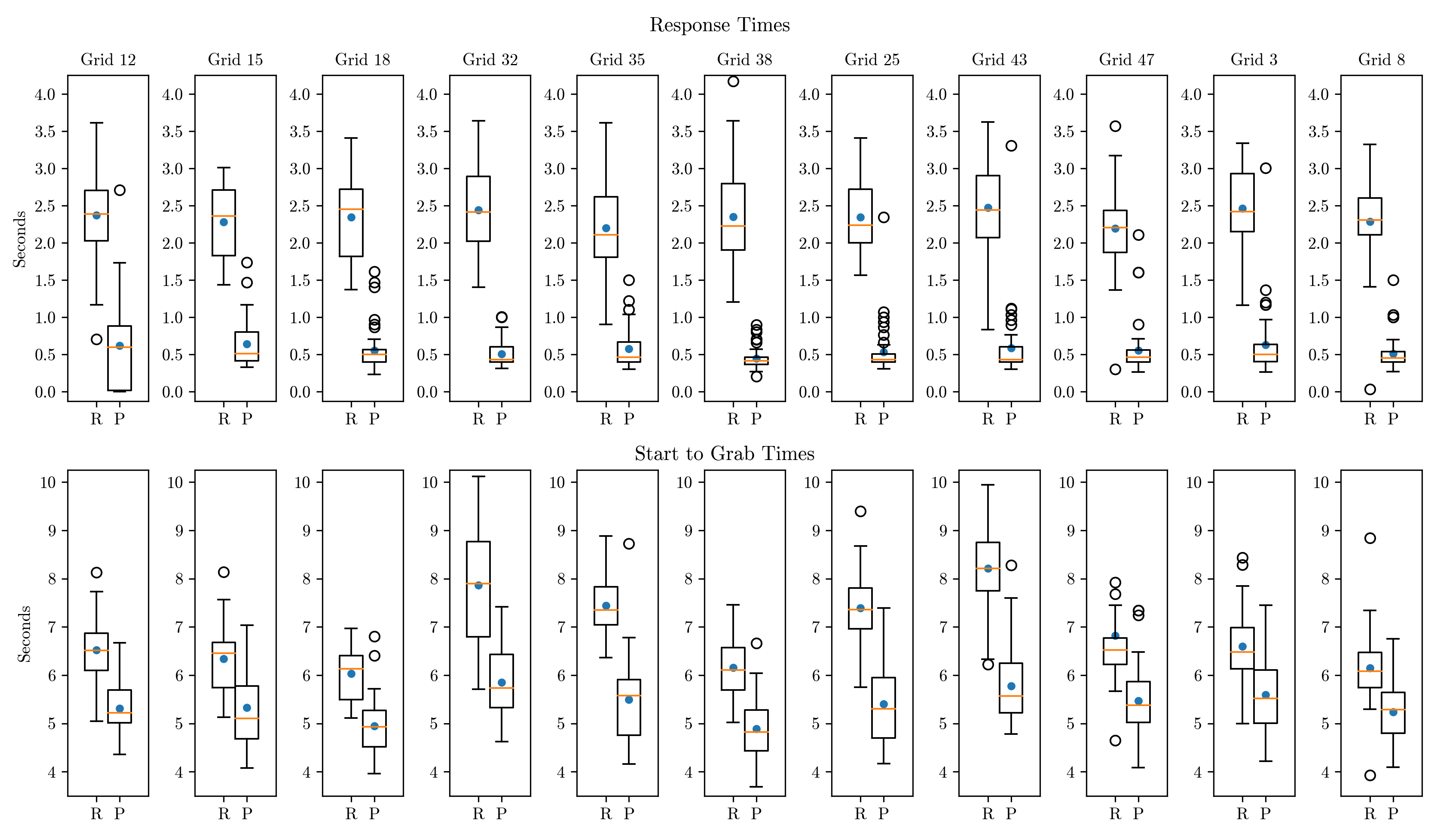}
  \caption{The first row and second row are the response and start-to-grab time boxplots, respectively. Each column corresponds to one of the eleven grid locations used in the experiment. Each subplot consists of two boxplots labeled R for reactive and P for preemptive. The yellow line corresponds to the median while the blue dot is the mean. As shown, there is a clear decrease in both response and start-to-grab times when employing preemptive control.}
  \label{fig:response_and_start_to_grab_time_boxplots}
\end{figure*}

\subsection{Experiment Setup}

We design an experiment where human participants sequentially place wooden cubes on specified locations on a $5 \times 10$ grid where each grid is identifiable by a numeric id. Eleven grid slots were chosen at random and used as the placement sequence for the experiment. These slots can be seen listed in Fig. \ref{fig:response_and_start_to_grab_time_boxplots}. Each grid consists of an area measuring $8 \times 8$cm resulting in a total rectangular workspace of $0.4 \times 1$m. Prior to the experiment, we recorded and labeled a set of 76 trajectories of variable length to use as the training data.

Each participant sits in front of a table with the grid and a Rethink Robotic's Sawyer 6DOF manipulator as shown in Fig. \ref{fig:motion_planning_example}. For perception, two Intel Realsense D435 cameras were used. One was mounted on the robot's base frame positioned to view and obtain human features while the second was mounted above pointing down towards the grid in order to detect the pose of the placed cubes. 
Fifteen participants were used in total. Each participant was informed that they would be working with the robot for six trials and that it would run two separate algorithms which would alternate between trials. All participants were told to abide by the following rules:
\begin{enumerate}
    \item Treat the object as if it was a full glass of water (\textit{to minimize rapid movements}).
    \item Keep the object occluded in their grasp until letting go (\textit{to keep final object position unknown until the end}).
    \item Do not start the trajectory with the held object occluded underneath the table or with the arm resting on the table.
    \item Do not place the next object down until the robot has returned to its ready pose.
    \item Try to place the object down with a similar speed for all six trials (\textit{in order to get the most unbiased results when comparing speed metrics}).
\end{enumerate}

\begin{figure*}
  \centering
  \includegraphics[width=\textwidth]{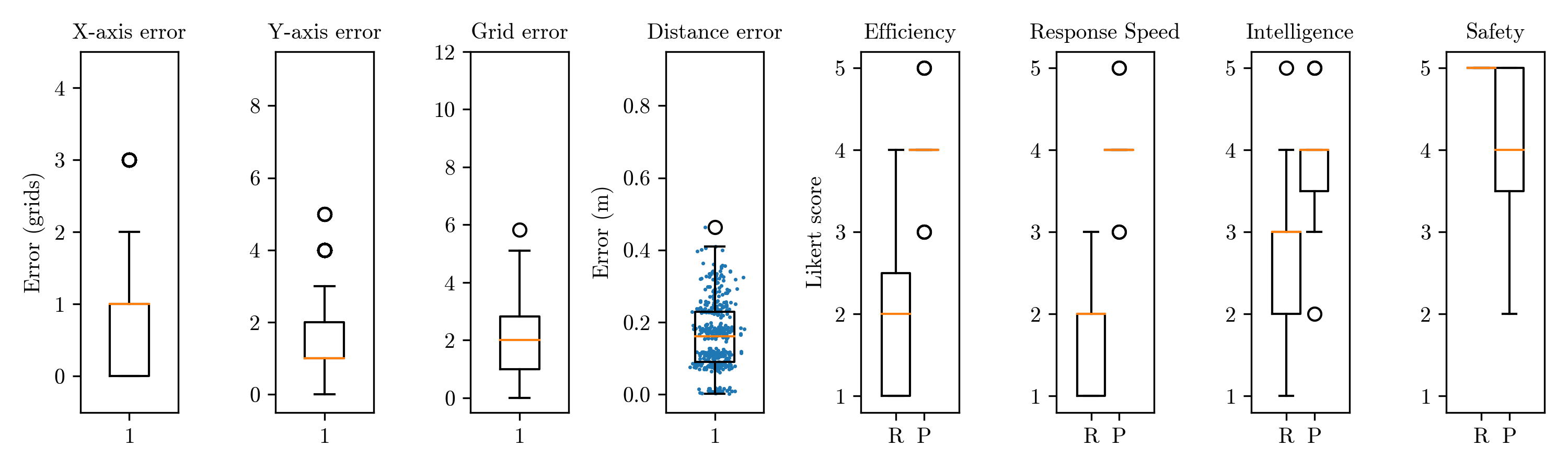}
  \caption{The first four boxplots showcase the prediction errors in terms of number of $x$-grids, number of $y$-grids, number of combined (euclidean) grids, and euclidean error in meters. The error in meters boxplot has all samples plotted with noise to show the distribution. 
  The last four boxplots showcase the Likert score results for the questionnaire from Sec. \ref{subsec:metrics}. Like Fig. \ref{fig:response_and_start_to_grab_time_boxplots}, these subplots each have two boxplots, R for reactive and P for preemptive.} 
  \label{fig:errors_and_questionnaire_boxplot}
\end{figure*}

\subsection{Metrics}
\label{subsec:metrics}
Two sets of metrics were recorded to evaluate the effectiveness of our preemptive control pipeline. The first were quantitative metrics measured by the program which include response time, start-to-grab time, and prediction error. Response time was measured by the time it took for the robot to start a motion plan starting from its ready pose. Start-to-grab time was measured by the time it took for the robot to actually grab the object starting from its ready pose. Prediction error was measured as both the number of grids the model was off by as well as the euclidean distance in meters.

The second set of metrics were responses to the following questionnaire provided by the participants at the end of the experiment:
\begin{enumerate}
    \item In terms of utilizing as much time as available, how efficient was each approach?
    \item How fast was the response time for each approach?
    \item How intelligent / intuitive did each approach seem?
    \item How safe did you feel working with each approach?
\end{enumerate}
Recorded responses consisted of a Likert score ranging from 1 to 5 for each approach for each question. Note that questions one (efficiency) and two (response time) were asked regardless of having access to quantitative measurements describing these two metrics. 
This was done in order to investigate how a human worker's perception of a robot coworker compared with measured results.


\section{EVALUATION}
\label{sec:evaluation}
\subsection{Quantitative Results}
The response and start-to-grab times for all grids and participants were boxplotted as shown in Fig. \ref{fig:response_and_start_to_grab_time_boxplots}. The average response time improvement was 1.78 seconds with all eleven grid positions having a $p$-value $<0.0001$ when comparing between their reactive and preemptive response times.

Still, the benefit of the reduced response times is largely dependent on the accuracy of our predictions. Here, the start-to-grab times can give a clearer view as to how much of the response time difference is actually used to move to an advantageous position. The average start-to-grab time improvement was 1.478 seconds which results in approximately 83$\%$ of the response time being utilized beneficially. 
Like response time, all eleven grid positions had a $p$-value $<0.0001$ when comparing between reactive and preemptive start-to-grab times. Therefore, we conclude that preemptive planning largely benefits the human-to-robot indirect placement transfer task.

Finally, the prediction errors can be seen in the left four boxplots of Fig. \ref{fig:errors_and_questionnaire_boxplot}. Note that the prediction errors plotted are the predictions used in the most recent predictive action before a definitive action was issued. Furthermore, as mentioned in Sec. \ref{subsec:motion_planning}, a tolerance of about 2.236 euclidean grids was used to minimize stop-and-go stuttering. Our average prediction error was 2.025 euclidean grids which is below this tolerance.

\subsection{Questionnaire Results}
We define the questionnaire metrics as efficiency, response speed, intelligence, and safety for each of the questions from Sec. \ref{subsec:metrics}. The questionnaire results for each of these metrics can be seen plotted in the right four boxplots of Fig. \ref{fig:errors_and_questionnaire_boxplot}. All questionnaire results had a $p$-value $<0.0001$ when comparing between the reactive and preemptive responses. As shown, most participants found the preemptive approach to be more efficient and intelligent with faster response speeds when compared to the reactive approach. This can most likely be attributed to the noticeable reduction in task completion time as well as the concurrent motion of the robot with the participant.

Unsurprisingly, the only metric that the preemptive approach scored worse on was safety where the reactive approach received unanimous perfect scores. Interestingly, a few participants (even after giving positive evaluations about task efficiency and stating that they did not feel like they were ever in danger) said they preferred the reactive robot as it was more comfortable to work with. This highlights the foreignness and unfamiliarity of working closely with robots that some may possess. From these results, a clear trade-off between task efficiency and human comfort seems to exist. Therefore, choosing between the two will likely heavily depend on the task at hand as well as the personal preference of the human worker.

\section{CONCLUSIONS}
\label{sec:conclusion}

In this paper, we introduced a novel preemptive control approach for human-to-robot indirect object placement transfers. This approach was shown to significantly reduce task completion time for indirect object placement transfers through a comprehensive case study between human participants and a robot manipulator. Furthermore, we concluded that preemptive robot behavior increases perceived efficiency and intelligence in exchange for general human comfort.  

Future work involves improving the performance of our model and planner, conducting more ablation studies, as well as expanding the problem space. 
Although our model was able to correctly predict the general placement location a majority of the time, predictions still had a large variance as shown in Fig. \ref{fig:errors_and_questionnaire_boxplot}. This can most likely be remedied by gathering more training data as well as possibly exploring non-recurrent-based architectures such as transformers.
Additional ablation studies would also provide better insight of the intent model. In particular, although we used previous work \cite{schydlo2018rnn_anticipation} to help design our feature vector, a comparison between using just gestures versus both gestures \& gaze would prove helpful.
Finally, we would like to expand our preemptive control work to harder indirect placement transfer problems such as those involving a human walking from afar.

\pagebreak









\bibliographystyle{ieeetr}
\bibliography{paper}

\end{document}